\documentclass[letterpaper, 10 pt, conference]{ieeeconf}  % Comment this line out if you need a4paper

\IEEEoverridecommandlockouts                              % This command is only needed if 
\usepackage{cite}
\usepackage{amsmath,amssymb,amsfonts}
\usepackage{algorithmic}
\usepackage{graphicx}
\usepackage{textcomp}
\usepackage{xcolor}
\usepackage{makecell, multirow, tabularx}
\usepackage{threeparttable}
\usepackage{multirow}
\usepackage{multirow}
\usepackage{multicol}
\usepackage{threeparttable}
\usepackage{colortbl}
\usepackage{xcolor}
\usepackage{soul}
\usepackage{booktabs}
\usepackage{comment}
\usepackage{graphicx}

\title{\LARGE \bf
Exploring Nutritional Impact on Alzheimer's Mortality: An Explainable AI Approach
}

\author{Ziming Liu$^{1}$, Longjian Liu$^{2}$, Robert E. Heidel$^{3}$ and Xiaopeng Zhao$^{4}$% <-this % stops a space
\thanks{$^{1}$Ziming Liu is with the Department of Mechanical, Aerospace and Biomedical Engineering, University of Tennessee, Knoxville, TN, 37916, USA {\tt\small zliu68@vols.utk.edu}}%
\thanks{$^{2}$Longjian Liu is Faculty with the Department of Epidemiology and Biostatistics, Dornsife School of Public Health, Drexel University, Philadelphia, PA, 19104, USA {\tt\small ll85@drexel.edu}}
\thanks{$^{3}$Robert E. Heidel is Faculty with the Department of Surgery, Graduate School of Medicine, University of Tennessee, Knoxville, TN, 37920, USA {\tt\small rheidel@utmck.edu}}
\thanks{$^{4}$Xiaopeng Zhao is Faculty with the with Department of Mechanical, Aerospace and Biomedical Engineering, University of Tennessee, Knoxville, TN, 37916, USA {\tt\small xzhao9@utk.edu }}}

\begin{document}

\maketitle
\thispagestyle{empty}
\pagestyle{empty}

%%%%%%%%%%%%%%%%%%%%%%%%%%%%%%%%%%%%%%%%%%%%%%%%%%%%%%%%%%%%%%%%%%%%
\begin{abstract}

In this study, we employ machine learning (ML) techniques along with explainable artificial intelligence (XAI) to delve into the intricate connection between nutritional status and mortality related to Alzheimer's disease (AD). To conduct the analysis, the Third National Health and Nutrition Examination Survey (NHANES III 1988 to 1994) and the NHANES III Mortality-Linked File (2019) databases are applied in this study. As a foundation for the XAI analysis, the random forest model is chosen as the primary model, and the Shapley Additive Explanations (SHAP) method is implemented to evaluate the significance of various nutritional features collected from blood testing. The findings of this study shed light on the crucial nutritional factors that impact AD and its associated mortality due to AD. Notably, Serum Vitamin B12 and age emerge as significant contributors. These results contribute to a more profound comprehension of the progression of AD and offer valuable insights into the influence of nutrition on the disease. The result demonstrates the potential of ML and XAI in uncovering complex relationships between nutritional status and AD, and providing interpretable explanations. Ultimately, this knowledge can inform interventions and strategies aimed at improving nutritional status and mitigating the mortality risks associated with Alzheimer's disease. 
\newline

\indent \textit{Clinical relevance}— This article investigates the link between nutrition and AD mortality using ML and XAI techniques with NHANES III. It identifies key nutritional factors and emphasizes their significance in managing AD and understanding disease progression.
\end{abstract}

%%%%%%%%%%%%%%%%%%%%%%%%%%%%%%%%%%%%%%%%%%%%%%%%%%%%%%%%%%%%%%%%%%%%%%%%%%%%%%%%
\section{INTRODUCTION}

%As a neurodegenerative disease, 
Alzheimer's disease (AD) has no effective cure today. Therefore, it is important to explore the effect of modifiable factors on disease progression and provide early treatments or interventions. According to recent studies, nutritional status has been demonstrated as a significant factor that represents a relationship with AD or dementia \cite{wang2004weight
%,sanders2018nutritional,sanders2016nutritional
}. In fact, malnutrition has been associated with faster disease progression and higher mortality rates \cite{soto2012weight
%,sanders2016nutritional
}. Moreover, the impact of disease progression may even be in the early stage of AD, such as mild cognitive impairment or mild dementia \cite{barberger2007dietary}. 

Not only AD but nutritional status has also been found to have a significant impact on various diseases, including heart disease and cancer. %Previous 
Studies have explored the relationship between nutrition intake and heart disease, noting associations between factors such as Vitamin E and C intake, as well as lipoprotein cholesterol density, with the risk of mortality caused by heart disease \cite{losonczy1996vitamin,steenland1998exposure}. Similarly, in older adult populations, there have been reports of a strong association between AD and cancer \cite{ospina2020association}. It is noteworthy that the interplay of different nutritional factors contributes to the synergy effects observed on various diseases, highlighting the complexity of these relationships.

Numerous machine learning (ML) techniques have emerged for the diagnosis of AD and demonstrate significant performance \cite{liu2022detecting
%,jiang2021memory,tanveer2020machine
}. However, the utilization of ML approaches provides accurate solutions but lacks transparency. Because the intricate association between nutritional status and AD remains unclear \cite{delgado2022nutritional}, compared to well-established biomarkers like neural imaging analysis, the predictive performance of ML in nutritional status-related AD diagnosis falls behind \cite{shah2013role}. Furthermore, it leads to limited applications of ML 
in assessing the impact of nutritional factors. 

Explainable artificial intelligence (XAI) utilizes statistical analysis to visualize and interpret machine learning frameworks' outcomes, bridging the gap between superior performances and a deep-level understanding of the pathology \cite{bordin2022explainable}. XAI has been applied in multiple studies ranging from neural imaging (Magnetic Resonance Imaging) \cite{kamal2021alzheimer}.
%,bordin2022explainable,shad2021exploring} 
Their adaptability could be applied in the area of neurodegeneration and assist in exploring the involvement of various regions in the determination of AD classification. 

The objective of this article is to investigate the intricate relationship between diverse nutritional statuses and mortality rates associated with AD by utilizing ML and XAI techniques. 
This analysis specifically focuses on comparing two aspects within groups: (1) mortality rates between AD and other diseases, and (2) mortality rates among AD, heart disease, and cancer. The goal is to gain insights into the specific impact of various nutritional factors on AD and discern any discrepancies compared to other prevalent diseases in older adults. To the best of the authors' knowledge, this is the first study to apply XAI in the nutritional analysis of AD.

\section{Materials and Methods}
\subsection{Database and Prepressing}
The Third National Health and Nutrition Examination Survey (NHANES III 1988 to 1994) \cite{test1}  and the NHANES III Mortality-Linked File (2019) \cite{test2} are applied. %in this study. 

The NHANES III was carried out by the National Center for Health Statistics (NCHS) of the Centers of Disease Control and Prevention in the US from October 1988 to October 1994, with the objective of collecting data on the health and nutritional status of the non-institutionalized US population aged two months and above. The study employed household interviews to gather demographic, medical history, and health behavior information, followed by physical exams and blood sample analysis in mobile examination centers. 
%Participants were required to fast for a specified duration to ensure the quality of the blood samples, and certified phlebotomists used standardized laboratory procedures during the collection process. 
To evaluate the dependencies of physical characteristics (Phi CHAR), including demographic and health-behavior information, on the classification performance using nutritional features, we define three types of feature lists: Phi CHAR, nutritional features, and nutritional features plus Phi CHAR. 93 different types of nutritional features from the blood sample analysis, as well as 12 Phi CHAR, are included in this study. Please see details of listed features in NHANES III 1988 to 1994 database\cite{test1}. Any missing data was assigned a value of -1.

For the NHANES III Mortality-Linked File, the NCHS collaborated with state Offices of Vital Statistics to link NHANES III data to death certificate records via the National Death Index, which is a centralized computerized index of death record information. This linkage offered a unique opportunity to conduct longitudinal analyses of baseline measurements and outcomes during the follow-up period, including survival or mortality. To examine the relationship between mortality caused by AD and survival, we used the Mortality-Linked File to track NHANES III participants until December 31, 2019. The study included n=8602 subjects, of whom n=208 subjects (AD) had died due to AD, and n=8394 subjects(Non-AD), were found the mortality caused by other diseases. Among n=8394 Non-AD subjects, n=2716 subjects(HD) and n=1826 subjects(CA) had died due to heart disease and cancer, respectively.

\subsection{Machine learning methodologies}

To assess the effectiveness of different machine learning approaches and identify the most suitable one as the base model for conducting XAI, we trained three models: a random forest, a support vector machine (SVM), and a two-layer fully connected neural network regression model. 

The two-layer fully connected neural network regression model employed a simple architecture with 32 artificial neurons in each layer and utilized the rectified linear unit (ReLU) activation function. In the case of the random forest model, we set the minimum leaf size to two(2) and employed 100 estimators across ten(10) random states. As for the SVM model, a radial basis kernel function was used.

\subsection{XAI analysis}
Shapley Additive Explanations (SHAP) is applied to assess the importance of input features in the classification model. SHAP provides comprehensive adjustments and enhances our understanding of the key nutritional features associated with the classification performance of Alzheimer's disease (AD). SHAP is a game-theoretic approach that systematically explains the output generated by any ML model. It combines local explanations with optimal credit allocation using classic Shapley values from game theory, as well as their relevant extensions \cite{lundberg2017unified}. SHAP offers both global and local explanations and can be applied to different ML and AI models, including artificial neural networks, decision trees, naive Bayes, and others. In the present study, SHAP is utilized as an indicator to assess the significance of nutritional features in the classification performance of AD.

%%%%%%%%%%%%%%%%%%%
\begin{figure}[htp]
\centering
\includegraphics[width=0.71\linewidth]{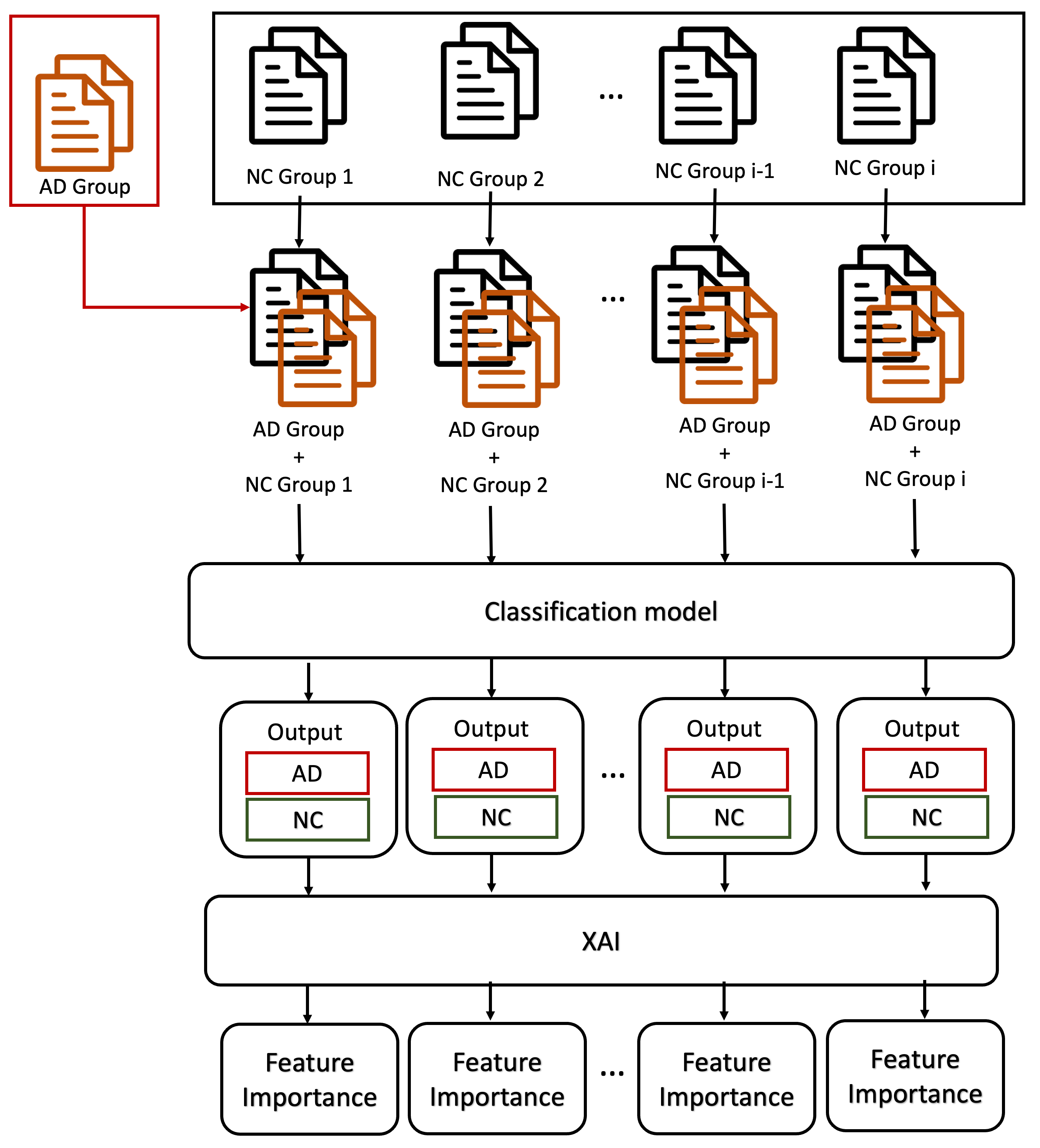}
\caption{Diagram of designed analysis approach (NC = Normal Control). \label{fig:diagram}}
\end{figure}
%%%%%%%%%%%%%%%%%%%

\subsection {Analysis Procedure}
Figure. \ref{fig:diagram} illustrates the methodology employed in this study. Given the imbalanced distribution between mortality caused by AD and survival, the normal control (NC) subjects were randomly divided into multiple groups, each comprising n=208 individuals, and combined with the n=208 AD subjects to create sub-datasets. These sub-datasets served as inputs for the ML methods to evaluate their diagnostic effectiveness in AD across all subgroups. To ensure comprehensive validation, a ten-fold cross-validation approach was employed for each system. The performance was assessed by calculating the average and standard deviation of scores from the Nx10 outputs, where N represents the number of normal control groups. Accuracy, precision, and recall scores were used to evaluate the model's performance.

The XAI approach proposed %in this study 
was applied to analyze the output of each model and generate feature importance weights for different inputs. These weights of importance were normalized using the softmax function. The average of the normalized weights was calculated and ranked in descending order to determine the importance of features.

Initially, the three proposed ML models were compared using the AD vs. Non-AD dataset as the input. The model with the highest performance was selected as the base model for subsequent XAI analysis. This XAI analysis was performed using three datasets trained by the selected based model: AD vs. Non-AD, AD vs. HD, and AD vs. CA, allowing for a comparison of the synergistic effects of nutritional status between different diseases and AD.

\section{Results}

Table \ref{tab: performance} presents the performance comparison results of the three proposed models. Based on the findings, the SVM model exhibited lower performance compared to the neural networks and random forests. Both neural networks and random forests demonstrated similar performance, but the random forest model exhibited relatively higher average recall scores. Therefore, the random forest model was selected as the base model for the subsequent XAI analysis.

%%%%%%%%%%%%%%%%
\begin{table}[htp] 
\centering
\caption{\label{tab: performance} Classification between AD and Non-AD mortality (mean (standard deviation) \%) using different ML methods.}
\begin{threeparttable}
\resizebox{\columnwidth}{!}{\begin{tabular}{@{}ccccccc@{}}\toprule

\textbf{ML method} & \textbf{Feature type} && \textbf{Accuracy} & \textbf{Precision} & \textbf{Recall}  \\ \midrule
\multirow{3}{*}{\textbf{Random Forest}} & Phi CHAR && 56.6(7.8) & 56.4(10.2) & 60.8(12.4)  \\ 
%\cmidrule{2-8}
 ~ & Nutritional && 70.7(7.2) & 79.9(11.7) & 65.5(11.5)  \\ 
%\cmidrule{2-8}
~ & Nutritional + Phi CHAR && 67.9(5.9) & 74.9(11.0) & 64.5(7.1) \\ 
\midrule

\multirow{3}{*}{\textbf{SVM}} & Phi CHAR && 52.9(6.3) & 57.4(8.8) & 52.2(3.2) \\ 
~ & Nutritional && 56.8(6.5) & 57.4(12.0)& 51.8(10.8)  \\ 
~ & Nutritional + Phi CHAR && 54.2(5.5) & 52.7(12.2) & 57.8(7.6) \\
\midrule

\multirow{3}{*}{\textbf{Neural Network}} & Phi CHAR  && 56.9(7.6) & 56.7(10.2) & 61.8(12.3) \\ 
~ & Nutritional && 71.2(7.1) & 82.1(11.7) & 54.1(11.2)   \\ 
~ & Nutritional + Phi CHAR && 68.1(5.7) & 76.2(11.3) & 53.3(7.9)\\
\bottomrule
\end{tabular}}
\end{threeparttable}
\end{table}
%%%%%%%%%%%%%%%%

The results displayed in Table \ref{tab: disease} demonstrate the classification performance across various databases, specifically for the comparisons of AD vs. Non-AD, AD vs. HD, and AD vs. CA. The findings indicate that the performance of the classification model was consistent across the three target datasets, yielding similar results. The results highlight the universality of using random forests for predicting the mortality of AD compared to other diseases. This incidental finding suggests that AD patients may exhibit a greater dominance of certain nutritional status features compared to individuals with other diseases.

%%%%%%%%%%%%%%%%
\begin{table}[htp] 
\centering
\caption{\label{tab: disease} Classification performance (mean (standard deviation) \%) on random forest with different type of datasets.}
\begin{threeparttable}
\resizebox{\columnwidth}{!}{\begin{tabular}{@{}ccccccc@{}}\toprule

\textbf{Dataset} & \textbf{Feature type} && \textbf{Accuracy} & \textbf{Precision} & \textbf{Recall}  \\ \midrule
\multirow{3}{*}{\textbf{AD vs. Non-AD}} & Phi CHAR && 56.6(7.8) & 56.4(10.2) & 60.8(12.4)  \\ 
%\cmidrule{2-8}
 ~ & Nutritional && 70.7(7.2) & 79.9(11.7) & 65.5(11.5)  \\ 
%\cmidrule{2-8}
~ & Nutritional + Phi CHAR && 67.9(5.9) & 74.9(11.0) & 64.5(7.1) \\ 
\midrule

\multirow{3}{*}{\textbf{AD vs. HD}} & Phi CHAR && 63.1(8.1) & 63.1(9.8) & 65.8(14.1) \\ 
~ & Nutritional && 72.3(7.3) & 80.9(10.7)& 58.6(10.8)  \\ 
~ & Nutritional + Phi CHAR && 72.6(6.0) & 79.4(10.4) & 62.5(8.4) \\
\midrule

\multirow{3}{*}{\textbf{AD vs. CA}} & Phi CHAR  && 65.0(7.1) & 64.5(10.3) & 67.3(11.4) \\ 
~ & Nutritional && 70.8(8.1) & 78.6(12.3) & 56.3(14.3)   \\ 
~ & Nutritional + Phi CHAR && 71.1(4.9) & 75.0(9.2) & 63.3(8.7)\\
\bottomrule
\end{tabular}}
\end{threeparttable}
\end{table}
%%%%%%%%%%%%%%%%

Table \ref{tab: adnonad}, Table \ref{tab: adhd}, and Table \ref{tab: adtu} present the top ten significant features obtained through SHAP analysis of trained random forest classification models using different input datasets (AD vs. Non-AD, AD vs. HD, and AD vs. CA) respectively. These results highlight that the features Serum vitamin B12 (VBP) and 3-Methylhistidine (mg) (NCPN3ME) consistently appear among the top ten rankings in all three analyses, focusing exclusively on nutritional features. Notably, VBP emerges as the most influential feature across all three models. However, when integrating Phi CHAR features into the models, age has demonstrated significant importance across the three analyses. Furthermore, in the AD vs. HD analysis, only Serum C-peptide (C1P), Red cell distribution width (RWP), and Serum total iron-binding capacity (TIP) appear in the top ten list, while in the AD vs. CA analysis, Urinary creatinine (URP), Serum folate (FOP), and Urinary cadmium (UDP) are present but not in the other two analyses.

%%%%%%%%%%%%%%%%
\begin{table}[htp] 
\centering
\caption{\label{tab: adnonad} Top 10 features and their average weights (\%) for AD vs. Non-AD classification.}
\begin{threeparttable}
\resizebox{\columnwidth}{!}{\begin{tabular}{@{}cccccc@{}}\toprule
\multicolumn{2}{c}{\textbf{Phi CHAR}}&\multicolumn{2}{c}{\textbf{Nutritional}}&\multicolumn{2}{c}{\textbf{Nutritional + Phi CHAR}}
\\ \midrule

\textbf{Feature} & \textbf{Weight} & \textbf{Feature} & \textbf{Weight} & \textbf{Feature} &  \textbf{Weight} \\ \midrule
AGE & 9.30 & VBP & 1.41& VBP &1.23\\
WAIST & 8.78 & GHP& 1.10 & AGE& 0.99\\
SMK & 8.50 & NCPNVK & 1.10& GHP & 0.98\\
WHR & 8.49 & BUP & 1.09 & NCPNVK & 0.97\\
BMI & 8.35 & UBP & 1.09& SBP & 0.96\\
SBP & 8.31 & SSBTP & 1.08 & SEP & 0.96\\
SEX & 8.30 & SEP & 1.08 & BUP & 0.96\\
DBP & 8.23 & UAP & 1.08 & UBP & 0.96\\
RACE & 8.07 & NCPN3ME & 1.08 & DBP & 0.96\\
EDU & 7.97 & BCP & 1.08 & UAP & 0.96\\
\bottomrule
\end{tabular}}
\begin{tablenotes} [para,flushleft]
\footnotesize 
Abbreviation: SMK = Smoke status, WHR = Waist-hip ratio, BMI = Boday mass index, SBP = Systolic blood pressure, DBP = Diastolic blood pressure, EDU = Education level, VBP = Serum vitamin B12 (pg/ml), GHP = Glycated hemoglobin: (\%), NCPNVK = NCC database Vitamin K (mcg), BUP = Serum blood urea nitrogen (mg/dL), UBP = Urinary albumin (ug/mL), SSBTP = Beta-trace protein (mg/L), SEP = Serum selenium (ng/mL), UAP = Serum uric acid (mg/dL), NCPN3ME = NCC database 3-Methylhistidine (mg), BCP = Serum beta carotene (ug/dL). Please see the full list of features in NHANES III database\cite{test1}. 
\end{tablenotes}   
\end{threeparttable}
\end{table}
%%%%%%%%%%%%%%%%

%%%%%%%%%%%%%%%%
\begin{table}[htp] 
\centering
\caption{\label{tab: adhd} Top 10 features and their average weights (\%) for AD vs. HD classification.}
\begin{threeparttable}
\resizebox{\columnwidth}{!}{\begin{tabular}{@{}cccccc@{}}\toprule
\multicolumn{2}{c}{\textbf{Phi CHAR}}&\multicolumn{2}{c}{\textbf{Nutritional}}&\multicolumn{2}{c}{\textbf{Nutritional + Phi CHAR}}
\\ \midrule

\textbf{Feature} & \textbf{Weight} & \textbf{Feature} & \textbf{Weight} & \textbf{Feature} &  \textbf{Weight} \\ \midrule
WAIST & 8.99 & VBP & 1.38 & VBP & 1.20\\
AGE & 8.91 & BUP& 1.10 & AGE& 0.98\\
SMK & 8.57 & UBP & 1.10 & DBP & 0.97\\
SEX & 8.56 & GHP & 1.09 & BUP& 0.97\\
DBP & 8.36 & C1P & 1.09 & UBP & 0.97\\
WHR & 8.35 & TIP & 1.09 & SBP & 0.97\\
SBP & 8.33 & RWP & 1.09 & GHP & 0.97\\
BMI & 8.26 & PBP & 1.09 & PBP & 0.96\\
RACE & 8.04 & NCPN3ME & 1.08 & C1P & 0.96\\
MAR & 7.93 & UAP & 1.08 & UAP & 0.96\\

\bottomrule
\end{tabular}}
\begin{tablenotes} [para,flushleft]
\footnotesize 
Abbreviation: SMK = Smoke status, DBP = Diastolic blood pressure, WHR = Waist-hip ratio, SBP = Systolic blood pressure, BMI = Body mass index, MAR = Marriage status, VBP = Serum vitamin B12 (pg/ml), BUP = Serum blood urea nitrogen (mg/dL), UBP = Urinary albumin (ug/mL), GHP = Glycated hemoglobin: (\%), C1P = Serum C-peptide (pmol/mL), TIP = Serum TIBC (ug/dL), RWP = Red cell distribution width (\%), PBP = Lead (ug/dL), NCPN3ME = NCC database 3-Methylhistidine (mg), UAP = Serum uric acid (mg/dL). Please see the full list of features in NHANES III database\cite{test1}. 
\end{tablenotes}   
\end{threeparttable}
\end{table}
%%%%%%%%%%%%%%%%

%%%%%%%%%%%%%%%%
\begin{table}[htp] 
\centering
\caption{\label{tab: adtu} Top 10 features and their average weights (\%) for AD vs. CA classification.}
\begin{threeparttable}
\resizebox{\columnwidth}{!}{\begin{tabular}{@{}cccccc@{}}\toprule
\multicolumn{2}{c}{\textbf{Phi CHAR}}&\multicolumn{2}{c}{\textbf{Nutritional}}&\multicolumn{2}{c}{\textbf{Nutritional + Phi CHAR}}
\\ \midrule

\textbf{Feature} & \textbf{Weight} & \textbf{Feature} & \textbf{Weight} & \textbf{Feature} &  \textbf{Weight} \\ \midrule
AGE & 9.11 & VBP & 1.37 & VBP & 1.16\\
SEX & 8.48 & SSBTP& 1.10 & AGE& 1.06\\
SMK & 8.32 & PBP & 1.09 & SEX & 0.99\\
WAIST & 8.30 & NCPN3ME & 1.08 & G1P & 0.97\\
BMI & 8.16 & URP & 1.08 & UDP & 0.97\\
DBP & 8.16 & SEP & 1.08 & GHP & 0.97\\
SBP & 8.16 & FOP & 1.08 & RWP & 0.96\\
WHR & 8.11 & FBP & 1.08 & BXP & 0.96\\
RACE & 8.04 & UDP & 1.08 & PBP & 0.96\\
MAR & 8.00 & NCPNVK & 1.08 & DBP & 0.96\\

\bottomrule
\end{tabular}}
\begin{tablenotes} [para,flushleft]
\footnotesize 
Abbreviation: SMK = Smoke status, BMI = Body mass index, DBP = Diastolic blood pressure, SBP = Systolic blood pressure, WHR = Waist-hip ratio, MAR = Marriage status, VBP = Serum vitamin B12 (pg/ml), SSBTP = Beta-trace protein (mg/L), PBP = Lead (ug/dL), NCPN3ME = NCC database 3-Methylhistidine (mg), URP = Urinary creatinine (mg/dL), SEP = Serum selenium (ng/mL), FOP = Serum folate (ng/mL), FBP = Plasma fibrinogen (mg/dL), UDP = Urinary cadmium (ng/mL), NCPNVK = NCC database Vitamin K (mcg), G1P = Plasma glucose (mg/dL), GHP = Glycated hemoglobin: (\%), RWP = Red cell distribution width (\%), BXP = Serum beta cryptoxanthin (ug/dL). Please see the full list of features in NHANES III database\cite{test1}. 
\end{tablenotes}    
\end{threeparttable}
\end{table}
%%%%%%%%%%%%%%%%

\section{Discussion and conclusion}
AD is a complex illness that is affected by the combined impact of various pathological factors. Hence, it is crucial to consider the diversity in AD diagnoses, particularly when examining the nutritional aspect, as the relationship between nutrition and AD is not well understood. Additionally, the interconnected influence of nutritional status on other diseases further complicates the matter. The notion that one disease has a greater influence than another is unfounded. Thus, this study introduces a new approach utilizing ML and XAI to enhance the comprehensibility of research on the nutritional status of AD from a holistic perspective.

In the proposed approach, Vitamin B12 plays a significant role in the model for detecting AD, regardless of its comparison with HD, CA, or Non-AD groups. Intriguingly, multiple studies report a link between B12 deficiency and increased risk for AD \cite{wynn1998danger}. Similar findings are observed for Serum C-peptide, which is only present in the list of AD vs. HD analysis but not in the other two groups. However, previous studies indicate that both AD and HD are influenced by Serum C-peptide \cite{frolich1998brain, li2015association}. Hence, we hypothesize that the proposed approach can effectively explore the varying strengths and weaknesses of dependencies in nutritional status analysis across different diseases.

In this study, due to the limited availability of AD cases, we employed a random sampling method to partition the normal control group and combine it with the AD group, resulting in a sub-dataset. However, this approach raises concerns regarding the potential skewness of the analysis, as the AD group remained unchanged across the sub-datasets. To address this, future studies will involve applying the proposed approach to a larger group in order to assess its performance more accurately.

Furthermore, the simple treatment of missing data in the study may have inadvertently diverted the model's attention from valuable features. To overcome this limitation, we plan to explore alternative imputation methods, such as Gated Recurrent Unit, to fill in the missing data. This will allow us to reevaluate the performance of the proposed approach and enhance its effectiveness in future investigations \cite{che2018recurrent}.

\bibliographystyle{ieeetr}
\bibliography{ref}

\end{document}